# A Comparison of Decision Analysis and Expert Rules for Sequential Diagnosis

Jayant Kalagnanam & Max Henrion

Department of Engineering and Public Policy,
Carnegie Mellon University,
Pittsburgh, PA 15213.
Tel: 412-268-2670;

## Abstract

There has long been debate about the relative merits of decision theoretic methods and heuristic rule-based approaches for reasoning under uncertainty. We report an experimental comparison of the performance of the two approaches to troubleshooting, specifically to test selection for fault diagnosis. We use as experimental testbed the problem of diagnosing motorcycle engines. The first approach employs heuristic test selection rules obtained from expert mechanics. We compare it with the optimal decision analytic algorithm for test selection which employs estimated component failure probabilities and test costs. The decision analytic algorithm was found to reduce the expected cost (i.e. time) to arrive at a diagnosis by an average of 14% relative to the expert rules. Sensitivity analysis shows the results are quite robust to inaccuracy in the probability and cost estimates. This difference suggests some interesting implications for knowledge acquisition.

## 1. Introduction

Although early work on automated diagnostic systems was much inspired by probabilistic inference and decision theory (Ledley & Lusted, 1959; Gorry & Barnett, 1968), many researchers later became disenchanted with this approach. Reasons cited include the difficulty of obtaining the required numerical probabilities and utilities, computational intractability, restrictive assumptions, and the apparent mismatch between the quantitative formalism of decision theory with human reasoning (Szolovits & Pauker, 1978). Thus, in the 1970s, this work was partly eclipsed by the development of AI approaches, which appeared more tractable and compatible with human thinking. More recently, however, there have been signs of renewed interest in the application of decision theoretic ideas in AI (Lemmer & Kanal, 1986; Horvitz, Breese & Henrion, in press). This has partly been due to increased misgivings about the assumptions and reliability of widely used heuristic methods for reasoning under certainty, and to the emergence of more tractable approaches based on probabilistic representations and decision analysis (e.g. Pearl, 1986; Henrion & Cooley, 1987).

Our focus here is on the application of decision analysis to troubleshooting, and particularly to sequential diagnosis, that is decisions about which diagnostic test to perform next and when to stop testing. We define the *test sequencing task* as the problem of finding the testing strategy that minimizes the total expected cost. This task is central in any kind of diagnosis, from medicine to mechanical and electronic devices.

If one has a causal model of the device that is entirely deterministic, logical analysis can identify possible explanations of the observed faults (e.g. Reggia et al, 1983; Genesereth, 1984; Milne, 1987). Intuitively it seems clear that test selection should involve some kind of balancing of the cost of the test against the chance it will provide useful information. This suggests that it is desirable to quantify the degree of belief in the logically possible explanations. Decision theory provides a way of quantifying the value of tests in the form of the *expected net value of information* (ENVI). This is a measure of the informativeness of the test, in terms of the expected value due to improved decision making, less the expected cost of performing the test.

Gorry & Barnett (1968) demonstrated the application of this approach to test sequencing in medical diagnosis. As they illustrate, the use of the ENVI can give one a voracious appetite for numbers. In

205

addition to the expected costs of the tests and quantifications of their diagnosticity, one requires the prior probabilities of the alternative hypotheses (diseases), and the expected total cost to the patient for each combination of treatment and disease (both correct and incorrect diagnoses). Further, the ENVI for multiple tests are not additive, so the expected value of two tests is not the sum of their expected values. Typically, the ENVI for each testing strategy must be computed separately. With many possible tests, the number of strategies is vast. For this reason, Gorry & Barnett used a "myopic" approach that considers only one test at a time, rather than an entire strategy. This is suboptimal in general, but they found it to be a good heuristic.

In many ways troubleshooting of mechanical and electronic devices is much easier than medical diagnosis. The usual end point is the positive identification of the faulty component to be repaired or replaced, so the possibility of ultimate misdiagnosis can often be neglected. If all testing strategies result in the same end point, the costs of treatment (or mistreatment) are irrelevant. Only the costs of performing the tests are of concern when selecting tests. Typically these costs may be quantified relatively easily in terms of the time required, that is the costs of the diagnostician's time and/or the machine downtime. Further simplifying matters is the fact that there often exists a causal model of the device which is largely or entirely deterministic. Such a luxury is not usually available to the medical diagnostician. Moreover, for many machine components and subsystems, tests can determine with virtual certaintly whether they are faulty. These aspects of troubleshooting make test selection far more amenable to a decision analytic approach than in medicine. In some cases, the optimal test selection algorithm can be very simple, as we shall see.

Our purpose in this work is twofold: First, to demonstrate the practicality of decision analysis for test sequencing for troubleshooting; and second, to compare experimentally the performance of this decision theoretic approach with a more conventional expert system approach employing test selection rules obtained from expert diagnosticians. After presenting the decision analytic formulation we will describe its application to the example of diagnosing of motorcycle engines, to examine its performance in terms of the expected cost (or time) to complete the diagnosis. Finally, we will examine the robustness of the results, and discuss their practical implications.

## 2. Decision analytic approach
We start by presenting a decision analytic approach to sequential diagnosis for a device representable by a deterministic fault model. We will not develop the formulation beyond the generality required for the experimental example.

### 2.1. Single level test sequence task
Suppose the system to be diagnosed consists of a set of elementary components, $i$ from $1$ to $n$. We assume that each component is in one of two states, working or faulty, and that the system fails in some observable way if there is a fault in any element. Given the system failure, we assume that there is a fault in exactly one component. This is a reasonable assumption for most systems, where the chance of two elementary components failing simultaneously is negligible. We assume that, given the system has failed, the probability of a component $i$ failing is $p_i$. Since the failures are mutually exclusive, the probabilities sum to unity.

We further assume that for each component there is a corresponding test which will tell us for certain whether it is working. The test might involve simple inspection, replacement of the component by an identical one known to work, probing with electronic test gear, or a variety of other operations. We assume that for each element $i$, the cost of testing is $c_i$, independent of the test sequence. We assume for simplicity that even if the fault is in the last element, which we can deduce if none others are faulty, we will test it anyway for positive identification.

The test sequence task is to find the strategy that minimizes the expected cost to identify the faulty component. It turns out that the optimal strategy is extremely simple: Select as the next element to test the one that has the smallest value of the ratio $c_i/p_i$, and continue testing until the faulty element is identified. We will call this the *C/P algorithm*.

206

It is not hard to prove this result. Let $S_{jk}$ be the strategy of testing each element i in sequence for $i$ from $1$ to $n$, until we find the faulty component. We certainly will have to incur the cost $C_1$ of testing the first component $e_1$. If that is working, we will have to test the second element, and so on. The probability that the *jth* element having to be tested is the probability that none of its predecessors failed, which is also the probability that either the jth element or one of its successors has failed, i.e. $\sum_{i=j}^{n} p_i$. Let $j$ and $k$ be two successive elements in this sequence, so $k=j+1$. Then the expected cost of strategy $S_{jk}$ is:

$$EC(S_{jk}) = C_1 + C_2 \sum_{i=2}^{n} p_i + ... + C_j \sum_{i=j}^{n} p_i + C_k \sum_{i=k}^{n} p_i + ... + C_n p_n \tag{1}$$

Let $S_{kj}$ be the same strategy, but with elements j and k exchanged. The expected cost of this strategy is:

$$EC(S_{kj}) = C_1 + C_2 \sum_{i=2}^{n} p_i + ... + C_k [\sum_{i=k}^{n} p_i + p_j] + C_j [\sum_{i=j}^{n} p_i - p_k] + ... C_n p_n$$

The difference in expected cost between these two strategies is then:

$$EC(S_{jk}) - EC(S_{kj}) = C_j p_k - C_k p_j$$

Assuming the probabilities are positive, we get:

$$EC(S_{jk}) > EC(S_{kj}) <=> C_j/p_j > C_k/p_k$$

In other words, strategy $S_{kj}$ is cheaper than $S_{jk}$ if and only if the *C/P* value for element *j* is greater than the *C/P* value for*k*. Thus any strategy with an element that has a higher C/P value than its successor can be improved upon by exchanging the successive elements. So for an optimal strategy, all elements must be in non-decreasing sequence of C/P value. This gives us our C/P algorithm.

This task is an example of what Simon & Kadane (1975) have called *satisficing search*. They show that a similar algorithm is optimal for the related search task, in which the events that each element contains a fault (or prize in their version) are independent rather than exclusive. They cite a number of similar results, and provide an algorithm for the generalized task with an arbitrary partial ordering constraint on the search sequence. See also Kadane & Simon (1977).

## 3. Experiment:
For the purposes of experimental comparison we chose the problem of diagnosing motorcycle engines. This choice was guided by two reasons. Firstly we had available an existing knowledge base for motorcycles. This allieviated the task of knowledge engineering. Secondly the motorcycle domain provided us with a problem that was small and simple enough to be manageable but still promising enough to make an interesting comparative study. The first goal was to explore the feasibility of implementing it in a real-world task domain, and obtaining the numerical probabilities and test costs required. The second goal was to evaluate its performance in terms of average cost to arrive at a diagnosis compared to the test sequence rules derived from human experts.

### 3.1. The rule-based approach:
Given the task domain, we identified 5 commonly occuring symptoms. Symptoms are understood as any deviations from expected behaviour. Symptoms can be directly observable or can be detected as a result of some measurement(s). Associated with each symptom is a set of elementary faults which might cause the given symptom. These elementary faults correspond to components that need to be repaired or replaced by the mechanic. The symptoms along with their corresponding symptoms are listed in table 3-1. Associated with each symptom is an *expert rule* that specifies the sequence of elements to be tested in that situation. These rules were obtained from extensive consultations with experienced mechanics and a standard reference manual (Harley Davidson). While eliciting these rules we asked the experts to keep in mind that the objective was to diagnose the fault as quickly as possible.

We interviewed three different mechanics. One of the mechanics refused to prescribe rules on the claim that all symptoms were quite straightfoward and with appropriate audio visual tests the faulty component could be exactly located. Unfortunately he was not able to characterize these tests beyond asserting that they are based on long years of experience. The other two experts did provide us with rules. Sometimes for a given symptom these rules differed markedly between experts. This difference did not affect our experimental study as we were only comparing the rules and C/p sequences for each given expert.

| Symptoms and Causes ||
|---|---|
| Symptoms | Corresponding Causes |
| poor-idling-due-to-carburettor | idle-speed-adjustment, clogged-speed-jet, air-leak-into-symptom, excess-fuel-from-accelerating-pump |
| starts-but-runs-irregularly | def-ignition-coil, def-ignition-module, improper-timing, air-cleaner/carburettor-adjustments, dirty-carburettor, engine-problems |
| charging-system-fails | stator-grounded, stator-defective, rotor-defective, def-regulator/rectifier |
| engine-turns-over-no-start-no-spark | air-gap-on-trigger-lobes, ignition-coil, circuit-between-battery-and-ignition-coil, ignition-module-defective |
| engine-turns-over-no-start-with-spark | spark-plugs, carburetion, advance-mechanism, improper-timing |

Table 3-1: Symptoms and Causes

### 3.2. The decision analytic approach:
For purposes of comparison we applied the decision analytic approach to the same set of symptoms we developed for the rule-based approach. We preserved all the assumptions made in section 2 for both approaches. Firstly all variables (namely symptoms and components) are binary, either working or not working. Both approaches also assume that there is only a single elementary component at fault.

The failure probabilities and test costs were obtained by interviews from motorcycle mechanics of many years' experience. The failure probabilities were assessed in groups, conditional on their common symptom. Given a symptom, the approximate absolute probabilities for different components (which cause the symptom) were assessed. The numbers were subsequently normalized. The costs were estimated as the average time in minutes the expert would take to determine whether each component was working.

The cost estimates between experts differed by upto a factor of three. The ordering of costs for a given symptom was consistent between experts. The difference in the absolute cost estimates can be attributed to personal differences. On the other hand, the probability estimates across experts was starkly different. For a given symptom, even the relative ordering of failure rates of components did not match across experts. One likely cause for this is the fact that different experts saw different samples of motorcycles. The mechanics who worked at dealerships were more likely to service new bikes which they had recently sold. Mechanics working in garages were more likely to see older bikes. As mentioned earlier these differences are not critical for this study.

For each symptom, we applied the C/P algorithm to obtain a test sequence, which we will refer to as the *C/P sequence* to distinguish it from the expert rule. We also showed the C/P sequences to the expert to check real world feasibility.

### 3.3. Results

Using the estimated cost and failure probabilities, we computed the expected cost in minutes for identifying the failed element for each symptom for both the expert and C/p sequences. Table 3-2 demonstrates the method used for comparison for a selected symptom for a specific expert.

| Selected Symptom: Poor - Idling due to Carburettor | | | | | |
|---|---|---|---|---|---|
| Components | Expert Rules | Cost(min) | Prob | C/p | C/p sequence |
| idle-speed-adjustments | 1 | 15 | .263 | 57 | 2 |
| clogged speed jet | 2 | 30 | .105 | 206 | 3 |
| air leak into system | 3 | 15 | .526 | 29 | 1 |
| excess fuel from accelerating pump | 4 | 30 | .105 | 206 | 4 |
| Expected Cost | | 50 | | | 32 |

**Table 3-2:** Method for a selected symptom

For each given symptom, the expected cost of diagnosis for both the expert rule and C/p sequence is tabulated in Table 3-3.

| Expected Cost for different experts | | | | |
|---|---|---|---|---|
| | Expert 1 | | Expert 2 | |
| | Expert Rule | C/p sequence | Expert Rule | C/p sequence |
| poor-idling due to carburettor | 8 | 8 | 50 | 32 |
| starts-but-runs-irregularly | 24 | 17 | 43 | 36 |
| charging-system-fails | 13.5 | 13.5 | 7 | 7 |
| engine-turns-no-start-no-spark | 17.5 | 16.6 | 55 | 53 |
| engine-turns-no-start-spark | 18.5 | 9 | 26.5 | 25.2 |
| Average Reduction | 16.5 | | 12 | |

**Table 3-3:** Expected Cost for different experts

The C/p algorithm provides a reduction in expected cost of diagnosis for 3 to 4 out of 5 cases, reducing the diagnostic time by an average (across experts) by 14%.

### 3.4. Sensitivity Analysis

The failure probabilities and test costs are of course quite approximate, being subjective judgments by the expert. An important question for any approach based on expert judgment, be it decision analytic or heuristic, is how much the results depend on the precise numbers used. Is it possible that the expert rules could in fact be optimal according to the C/P algorithm, if only we had obtained the expert's "actual" probabilities and costs? Let us examine how much bias or error would there have to be in the assessment process for this to happen.

For a given symptom, we associate with each component a probability distribution over the failure rate and the cost of testing. Since the failure rate itself is a probability it's range is [0,1]. Therefore we describe the failure rate as a logoddnormal distribution with the mean as the estimated value $p$ and an error factor $s$. The range $[m/s, m.s]$ encloses 70% of the probability distribution. Similarly we associate a lognormal distribution for the cost with mean as the estimated value $C$ and an error factor $s$. Now the *difference* between the expected cost for C/p sequence and the expert rule is itself a probability distribution with $s$ as a parameter. We assume that $s$ is the same for *all* the distributions. Figure 3-1

209

shows the cumulative density function of the *difference* (of the expected cost between the C/p sequence and expert rule) for a selected symptom with $s=2$. Figure 3-1 also graphs the boundaries for *difference* which encloses 70% of the distribution around the mean. From this graph we see that for $s \sim 2.5$, the lower boundary intersects the zero-line. In other words if the error factor is 2.5 then the expert rule and the C/p sequence have the same expected cost.

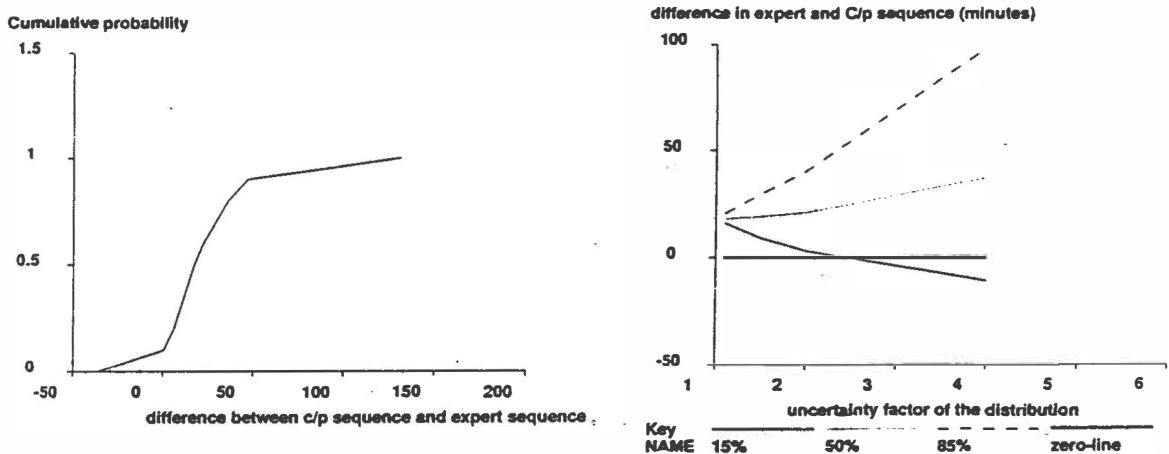

**Figure 3-1:** Sensitivity Analysis for "poor-idling'

Such an analysis for all symptoms, over all experts shows that the C/p sequence dominates the expert rule for error factors up to 2.5. This suggests that the results obtained are fairly robust to errors in cost and failure rate estimates.

### 3.5. Further consulations with experts
A major motivation for constructing formal models is the possibility that such models may improve on the intuitive inference of the expert. Hence when the formal models lead to results different from those suggested by experts, it becomes critical to explain these differences to the experts. If explanations (from within the formal framework) are acceptable and the results insightful then the formal models have served their purpose.

Keeping the above criterion in mind, we went back to the experts with the C/p sequences. Firstly, both experts found the C/p sequences to be feasible. Hence the C/p sequences did not violate any implicit realworld constraints. When it came to convincing the experts about the superiority of the C/p sequence the results were mixed. Expert 2 was a little surprised at the difference between his sequencing and the C/p sequencing. A careful explanation of the C/p algorithm convinced him of the dominance of the C/p sequence. He readily agreed that the C/p sequence was better at minimizing the expected cost. On the other hand expert 1 was not impressed by the C/p sequence. He felt that his sequences followed a causal pathway while checking for faults which he thought was more desirable. This bias for causal paths made him reject the C/p sequences.

## 4. Discussion
The results presented in section 3.3 suggest that the expected cost of diagnosis is significantly lower (at least for some cases) for the decision analytic approach. This result can be understood in light of the difference in test sequences obtained from human experts relative to those derived from the C/p algorithm. In order to explain and draw conclusions from this result we need to discuss three possible shortcomings of the experimental setup.

The cost and failure rate estimates are approximate. Therefore it is possible that the inaccuracies in these estimates might affect the test sequences derived from the C/p algorithm. However the sensitivity analysis in section 3.4 shows that the results are quite robust to inaccuracies in the cost and failure rate estimates.

The decision analytic approach sequences tests to optimize the expected cost of diagnosis. This approach is in fact too simple to model other constraints on test sequences which might arise from the shape and structure of the machine under consideration. As a result the C/p algorithm might in fact be ignoring implicit but vital constraints on the test sequences. On the other hand the expert rule might have implicitly accounted for such constraints and the difference in the two approaches might be attributed to this factor. In order to check for this possibility, we presented the experts in question with the C/p sequence for their comments. Specifically we were interested in the real world feasibility of the C/p sequences. We encouraged the experts to suggest reasons as to why the expert sequences might be preferable, so as to identify any implicit real world constraints that might be violated by the C/p sequence. The C/p sequences used are ones that were found acceptable to the experts. We interpret this to mean that the C/p sequences are indeed feasible ones.

The last question we need to worry about is whether the human experts (in the task domain) were actually attempting to minimize the expected time for diagnosis. There is a real possibility that the experts in fact had some other objective function. It is indeed more than plausible that experts have over time evolved these sequences to maximize economic return for each given symptom. The rules might maximize the time for diagnosis, subject to constraint that customer will not think it unreasonable.

## 5. Conclusions

The results of this study clearly indicate that the test sequences provide by the experts (in the task domain) are suboptimal. Unfortunately there is uncertainty regarding the objectives which motivate the expert test sequences. This restrains us from drawing firm conclusions about the efficacy of human intuition for this task domain. But it is important to remember that one of the experts accepted the validity of the C/p sequence and felt that the results are likely to be of practical interest. This suggests that normative theories of decision making are capable of obtaining results which go beyond current expert opinion.

This study also provides some valuable insights regarding knowledge acquistion for diagnostic expert systems. Diagnostic problem solving can be understood as follows:

- Given a symptom we have a set of components which can potentially explain the symptom. The physics and the structure of the problem provides a partial order on the set of test sequences allowable on the set of components. Since the partial order does not completely constrain the test sequence, we can optimize the expected cost of diagnosis over this set of feasible test sequences.

The knowledge required for this optimization could be acquired in two ways:

1. As an expert rule which picks one test sequence from this set of feasible test sequences. This corresponds to the rule-based approach of our study.

2. We can push level of knowledge acquistion a level deeper and explicitly represent the cost and failure rates. These are once again assessed from experts. This corresponds to the decision analytic approach of our study.

Our study suggests that it possible to explicitly represent the cost and failure rates and it provides better control on the objective of optimizing the expected cost of diagnosis.

## Acknowledgements

This work was supported by the National Science Foundation, under grant IST 8603493. We are much indebted to David Keeler, Woody Hepner and Larry Dennis for lending us their expertise as motorcycle mechanics. We are grateful to Jeff Pepper and Gary Kahn of the Carnegie Group, Inc, for making available the expert system, and to Peter Spirtes and J. Kadane for valuable suggestions.